\documentclass[letterpaper, 10 pt, conference]{ieeeconf}  
\IEEEoverridecommandlockouts                              

\usepackage{graphicx}
\usepackage{booktabs}
\usepackage{hyperref}
\usepackage[table]{xcolor}
\usepackage{orcidlink}
\usepackage{multirow}
\usepackage{colortbl}
\usepackage{amssymb}
\usepackage{amsmath}
\usepackage{tabularx}
\usepackage{pifont}
\newcommand{\xmark}{\ding{55}}
\newcommand{\cmark}{\ding{51}}
\usepackage[caption=false,font=footnotesize]{subfig}
\title{\LARGE \bf
PrismAD: Decoupled Planning via Semantic Mixture-of-Planners for End-to-End Autonomous Driving
}

\begin{document}
\author{Kang Ding$^{1,2,3}$, Zhigui Lin$^{5}$, Hongsong Wang$^{4}$, Jie Gui$^{3}$, Qi Liu$^{1,2,4}$,\\
Zhe Wang$^{6}$, Luqi Tang$^{6}$, and Lei He$^{1,2\dagger}$%
\thanks{*This work is supported by National Key R\&D Program of China 2024YFB2505500 and Guangxi Science and Technology Major Program (No.AA24206054)}
\thanks{$^{\dagger}$Corresponding Author: Lei He. E-mail: helei2023@tsinghua.edu.cn}
\thanks{$^{1}$School of Vehicle and Mobility, Tsinghua University, Beijing 100084, China.}%
\thanks{$^{2}$State Key Laboratory of Intelligent Green Vehicle and Mobility, Tsinghua University, Beijing 100084, China.}
\thanks{$^{3}$School of Cyberspace Security, Southeast University, Nanjing 210096, China.}
\thanks{$^{4}$School of Computer Science and Engineering, Southeast University, Nanjing 210096, China.}%
\thanks{$^{5}$School of Automotive Engineering, Wuhan University of Technology, Hubei Key Laboratory of Advanced Technology for Automotive Components, Hubei Collaborative Innovation Center for Automotive Components Technology, and Hubei Research Center for New Energy \& Intelligent Connected Vehicles, Wuhan 430070, China.}%
\thanks{$^{6}$SAIC GM Wuling Automobile Co., Ltd., Guangxi Laboratory of New Energy Automobile, and Guangxi Key Laboratory of Automobile Artificial Intelligence, Liuzhou 545000, China.}%
}

\maketitle
\thispagestyle{empty}
\pagestyle{empty}

\begin{abstract}
This letter presents PrismAD, a decoupled end-to-end autonomous driving framework based on a Semantic Mixture-of-Planners. Existing planners usually aggregate heterogeneous scene tokens into a coupled representation space, forcing a single planning branch to jointly model agent interaction, road geometry, and driving intention. Such coupling may weaken factor-specific reasoning and obscure the contribution of different planning cues. To address this limitation, PrismAD partitions scene tokens into interaction, geometry, and intent groups, and assigns them to independent planning experts with the same architecture but separate parameters. Each expert learns a specialized motion-planning representation, while a semantics-aware router adaptively aggregates expert predictions with separate routing weights for motion prediction and ego planning. Sparse top-$K$ activation with noisy gating is further introduced to improve routing robustness and reduce unnecessary expert computation. Extensive experiments on the nuScenes open-loop dataset and NeuroNCAP closed-loop benchmark demonstrate that PrismAD exhibits competitive performance. Our code will be released soon.
\end{abstract}

\section{INTRODUCTION}

End-to-end autonomous driving has become an important research direction because it directly maps sensory observations to planning decisions and reduces the hand-crafted dependencies in traditional modular pipelines. Recent planning-oriented methods have made steady progress by integrating perception, prediction, mapping, and planning into unified neural frameworks~\cite{hu2023_uniad,Jiang2023VADVS,Sun2024SparseDriveEA}. In particular, query-based and sparse-representation planners have shown promising efficiency by representing traffic scenes with compact object, map, and ego tokens. However, planning remains challenging because a safe ego trajectory is jointly determined by multiple heterogeneous factors, including dynamic interactions with surrounding agents, local road geometry, and high-level navigation intention.
Most existing end-to-end planners\cite{Zheng2024GenADGE,Chi2025ImpromptuVO,Song2025DontST,Zhang2025BridgingPA} aggregate these heterogeneous scene tokens into a single coupled representation space. Although this unified design simplifies the model architecture, it forces one planner to simultaneously reason about factors with different semantic roles. For example, surrounding-agent tokens mainly support interaction reasoning and collision avoidance, map tokens provide geometric constraints from lanes and road boundaries, and command-related tokens determine the intended driving direction. Treating all these factors as a single undifferentiated token set can obscure their individual contributions to the final decision and may weaken both specialization and interpretability. This limitation becomes more pronounced in complex driving scenarios, where different factors dominate the planning decision under different contexts.
\begin{figure}[tbp]    
    \centering    
    \includegraphics[width=0.48\textwidth]{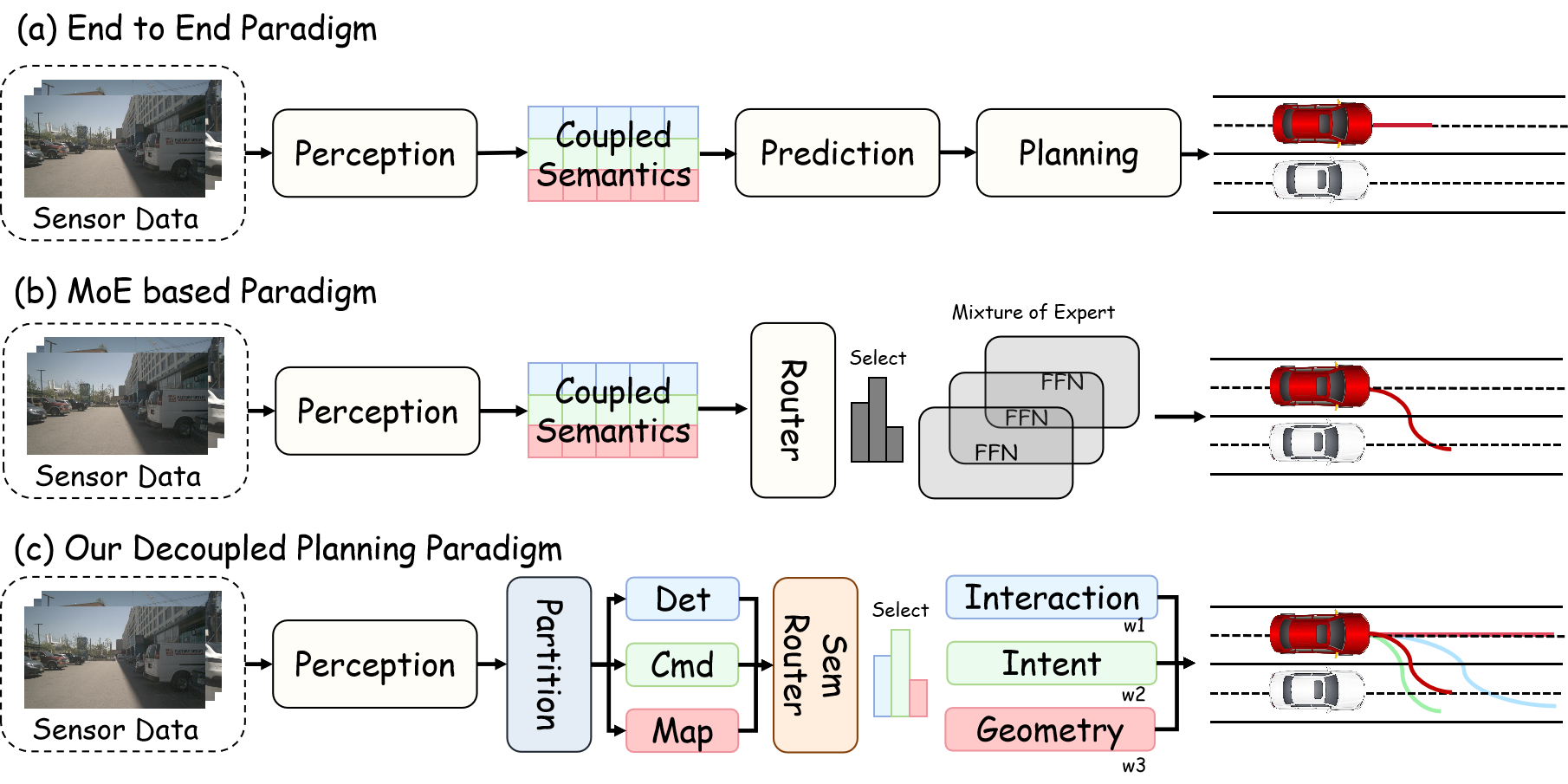}    
    \caption{\textbf{Comparison of our approach with existing methods.} (a) Traditional end-to-end Paradigm. (b) Classical MoE-based Paradigm. (c) Our proposed Paradigm.}    
    \label{fig:intro}
    \vspace{-3mm}
\end{figure}
A natural solution is to introduce expert specialization. Recent studies have explored Mixture-of-Experts for autonomous driving in diverse forms, including autoregressive trajectory generation, task-adaptive perception and prediction, vision-language-action planning, cooperative multi-agent fusion, scene-adaptive routing, and reward-balanced motion planning~\cite{feng2025artemis,jiang2026expertad,yang2025drivemoe,song2026unimm,you2026samoe,sun2025generalizing}. These methods demonstrate the value of expert specialization for handling heterogeneous scenarios, reducing task interference, and improving conditional computation. However, their expert designs are mostly organized around scenario distribution, task demand, motion pattern, camera view, driving skill, or reward objective, rather than explicitly decomposing the semantic factors that determine an ego trajectory. In autonomous driving, planning is jointly shaped by surrounding-agent interaction, road geometry, and navigation intention. A planner should therefore select experts for different inputs and learn factor-level representations corresponding to these semantic cues. Moreover, motion prediction and ego planning may require different expert preferences, since the former depends more on agent interaction while the latter is more constrained by road geometry and high-level intention. Therefore, a planning-level mixture mechanism with semantic routing is needed.

To address these issues, we propose PrismAD, a decoupled planning framework based on a Semantic Mixture-of-Planners. Instead of processing all scene tokens in a coupled space, PrismAD partitions perception tokens into interaction, geometry, and intent groups. Interaction tokens are selected from dynamic object queries to model ego-agent and agent-agent relations, geometry tokens are selected from map queries to encode road topology and structural constraints, and intent tokens are constructed from navigation-command embeddings to capture high-level planning intention. Each group is concatenated with an ego token and fed into an independent planning expert with the same architecture but separate parameters.

PrismAD further introduces a semantics-aware router to adaptively aggregate expert predictions. The router estimates expert importance from ego-centric queries and scene tokens, and produces separate routing weights for motion prediction and ego planning. During inference, sparse top-$k$ activation selectively activates scenario-relevant experts while avoiding unnecessary computation. The interaction expert remains active as the basic branch, while the geometry and intent experts are selected according to routing scores.

The main contributions of this letter are summarized as follows. First, we propose a semantic token partition strategy that explicitly decouples scene tokens into interaction, geometry, and intent groups, enabling factor-aware planning representation learning. Second, we design a Semantic Mixture-of-Planners, where each expert is a complete motion-planning branch rather than a generic expert inserted into an intermediate network layer. Third, we introduce a semantics-aware router that produces separate expert weights for motion and planning, supporting adaptive fusion and sparse expert activation. Finally, extensive experiments on nuScenes, NeuroNCAP, and Turning-nuScenes demonstrate that PrismAD consistently improves strong planning baselines and achieves competitive performance in both open-loop and closed-loop planning evaluations.

\section{METHOD}
\begin{figure*}[htbp]
    \vspace{2mm}
    \centering        
    \includegraphics[width=0.92\textwidth]{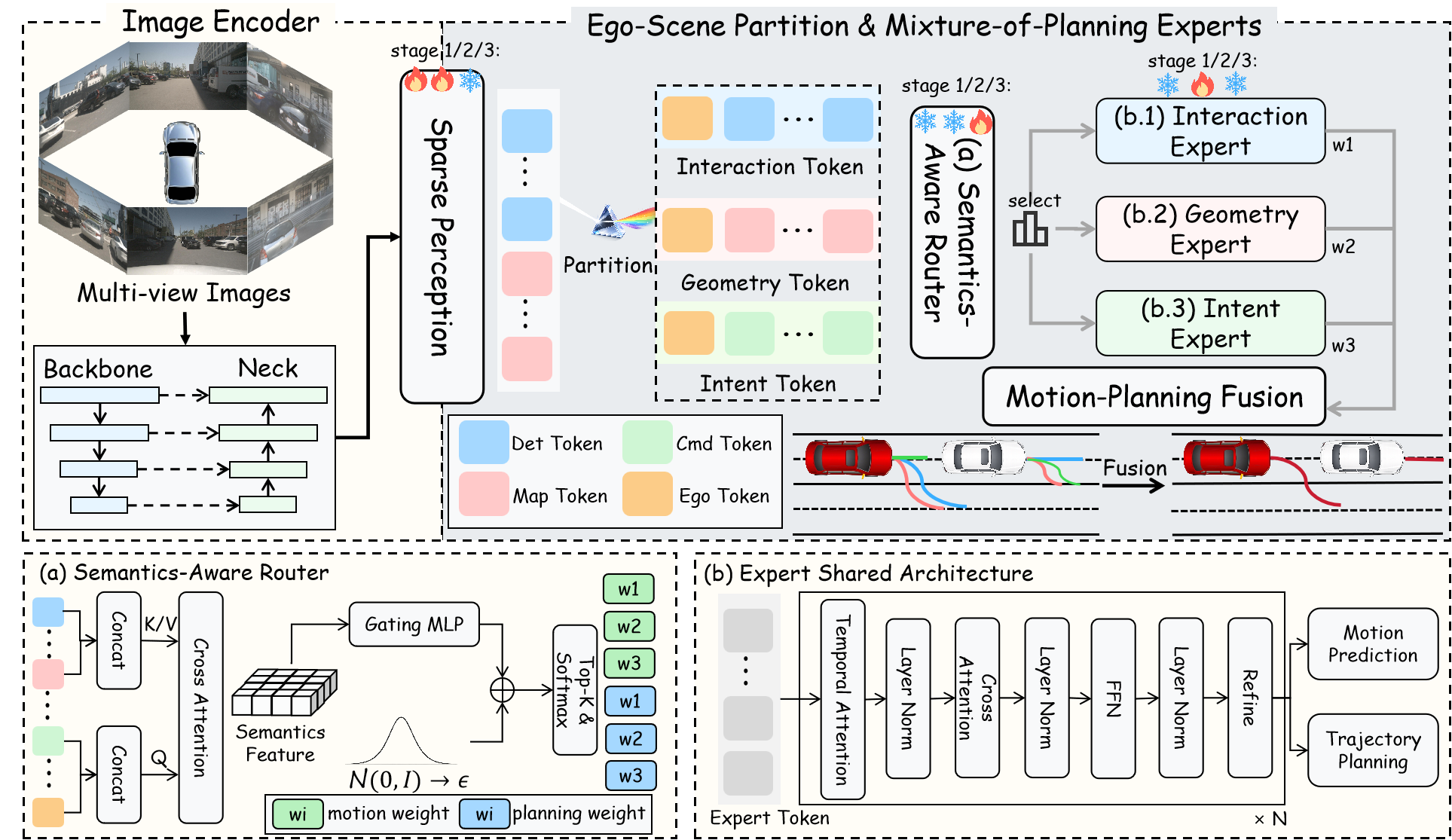}        
    \caption{
    Architecture of PrismAD. The perception backbone extracts sparse scene tokens from multi-view images, which are partitioned into interaction, geometry, and intent groups. Each group is processed by an independent planning expert, while a semantics-aware router estimates motion and planning weights for sparse expert selection and final trajectory fusion.
}  
    \label{fig:overview}
\end{figure*}

Existing end-to-end planners usually aggregate heterogeneous scene tokens into a unified representation space, which weakens factor-specific reasoning and limits interpretability. To address this issue, PrismAD decouples scene tokens through semantic partitioning and models different planning factors with independent experts. A semantics-aware router further performs adaptive expert aggregation for final motion prediction and ego planning. The overall architecture is shown in Fig.~\ref{fig:overview}.
\begin{figure}[tbp]
    \centering
    \vspace*{2mm}
    \begin{minipage}{0.48\columnwidth}
        \centering
        \includegraphics[width=\linewidth]{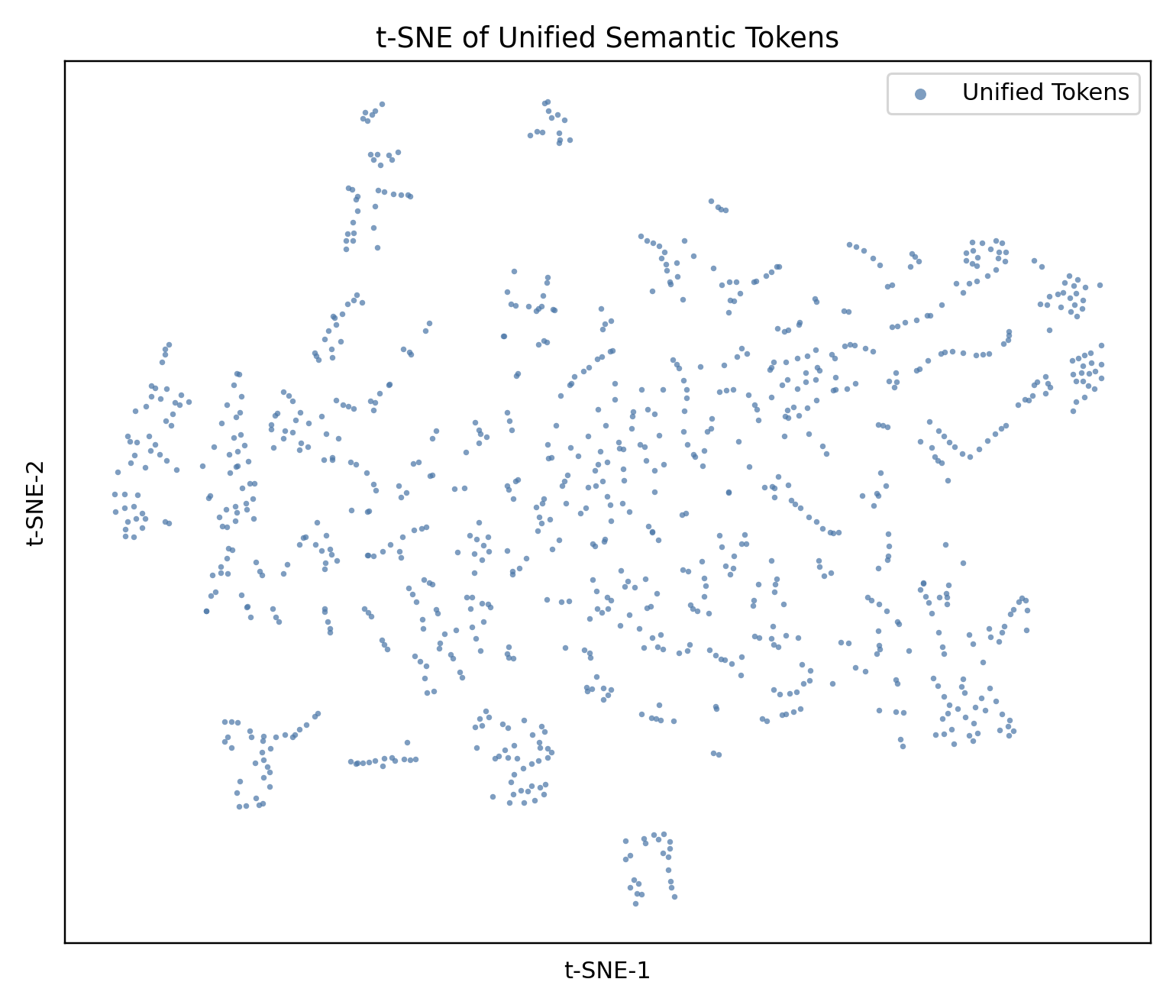}
    \end{minipage}
    \hfill
    \begin{minipage}{0.48\columnwidth}
        \centering
        \includegraphics[width=\linewidth]{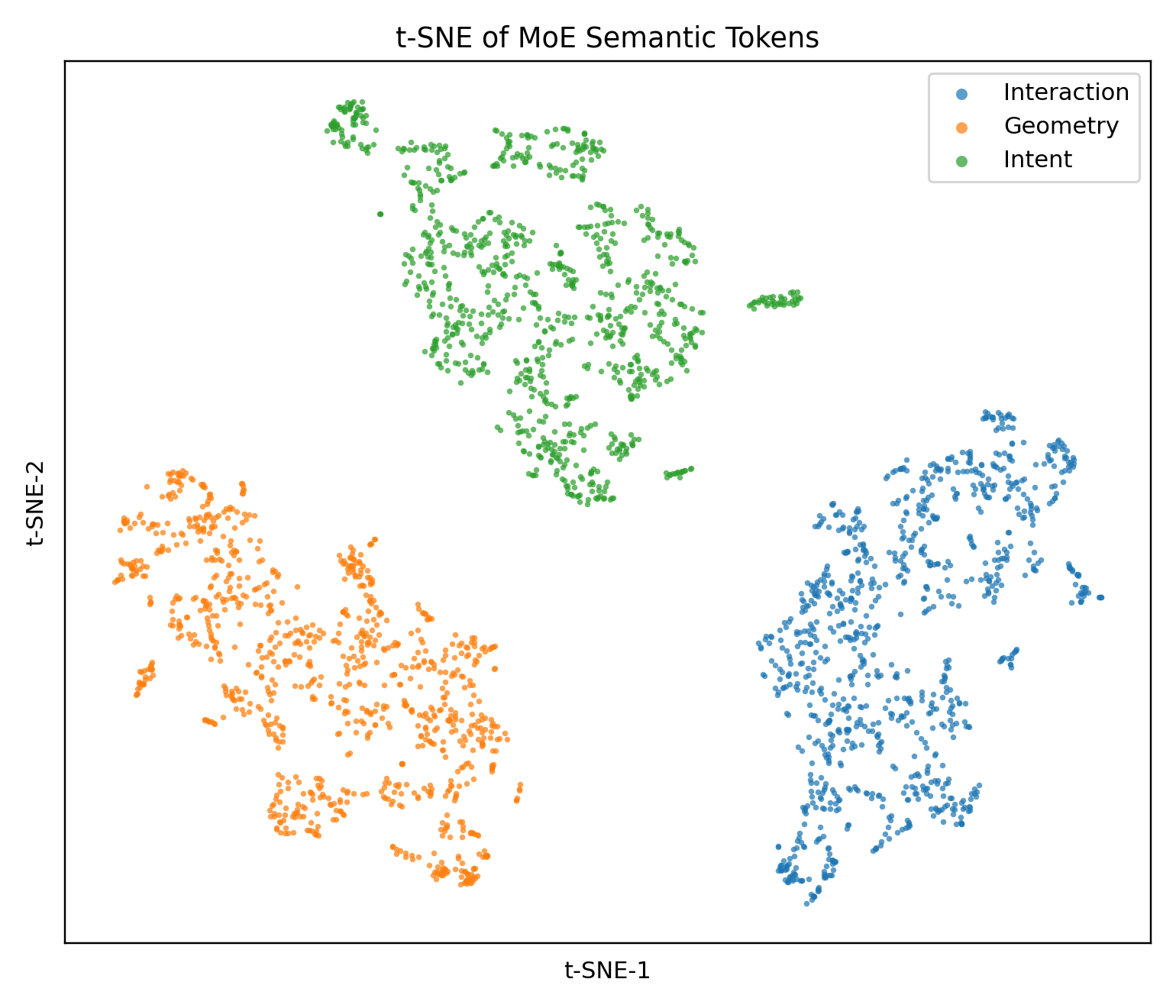}
    \end{minipage}
    \caption{t-SNE visualization of interaction, geometry, and intent ego tokens on the nuScenes, showing distinguishable feature distributions induced by semantic token partition.}
    \label{fig:tsne}
\end{figure}

\subsection{Semantic Token Partition}

Autonomous driving planning is jointly determined by multiple factors, including interactions with surrounding agents, road geometry, and high-level driving intention. However, existing end-to-end planners usually concatenate all scene tokens and process them in a coupled representation space. Such a design forces one planner to simultaneously model factors with different semantic meanings, which may obscure their individual contributions to the final decision.

To explicitly decouple these factors, we introduce a semantic token partition strategy. Given the perception outputs from the perception module, we divide the scene tokens into three semantically distinct groups: interaction tokens, geometry tokens, and intent tokens. The interaction tokens are selected from high-confidence dynamic object queries, which mainly encode the motion states and potential interactions of surrounding agents. The geometry tokens are selected from map element queries that describe the local road structure, lane boundaries, and drivable regions. The intent tokens are constructed from command embeddings and ego status tokens, capturing the high-level navigation instruction and the current state of the ego vehicle.

Formally, let $\mathcal{T}$ denote the set of scene tokens produced by the perception module. We partition it into three subsets:
\begin{equation}    
\mathcal{T} = \{\mathcal{T}_{\mathrm{inter}}, \mathcal{T}_{\mathrm{geo}}, \mathcal{T}_{\mathrm{intent}}\},
\end{equation}
where $\mathcal{T}_{\mathrm{inter}}$ represents interaction-aware tokens, $\mathcal{T}_{\mathrm{geo}}$ represents geometry-aware tokens, and $\mathcal{T}_{\mathrm{intent}}$ represents intent-aware tokens. Each subset is further concatenated with an ego token, which provides a common ego-centric reference for all experts:
\begin{equation}    
\mathbf{X}_{e} = [\mathcal{T}_{e}; \mathbf{x}_{\mathrm{ego}}],     
\quad e \in \{\mathrm{inter}, \mathrm{geo}, \mathrm{intent}\}.
\end{equation}

This partitioning converts the coupled planning representation into semantically aligned expert inputs. As a result, each expert receives tokens that are closely related to its corresponding planning factor, improving both specialization and interpretability. To qualitatively examine this design, we visualize the ego-token features of the three semantic branches using t-SNE in Fig.~\ref{fig:tsne}. The interaction, geometry, and intent tokens exhibit clearly distinguishable distributions in the feature space which provides qualitative evidence that our proposed partition encourages different semantic branches to form distinguishable feature distributions, and supports the use of specialized planning experts.
\subsection{Decoupled Expert Representation Learning}
After semantic partitioning, we further employ multiple planning experts to learn factor-level planning representations. As shown in Fig.~\ref{fig:overview} (b), the experts share the same network architecture but do not share parameters. This design keeps the model structure consistent across different semantic branches while allowing each expert to learn an independent representation space.

Given the input $\mathbf{X}_{e}$, the expert first incorporates temporal information from historical ego and instance features through temporal attention. It then performs intra-branch token interaction to model the relationships among tokens within the same semantic group. The resulting representation is refined by feed-forward layers and finally decoded into motion and planning predictions.

For the $e$-th expert, the representation learning process can be defined as follows:
\begin{equation}
    \mathbf{H}_{e} = \Phi_{e}(\mathbf{X}_{e}, \mathbf{X}_{e}^{\mathrm{hist}}),
\end{equation}
where $\Phi_{e}$ denotes the expert network and $\mathbf{X}_{e}^{\mathrm{hist}}$ denotes the temporal features. Since the parameters of $\Phi_{e}$ are independent across experts, each branch can specialize in a different planning factor. The interaction expert focuses on agent-agent and ego-agent relations, the geometry expert emphasizes road topology and map constraints, and the intent expert models command-conditioned planning behavior.

Each expert predicts both agent motion and ego-vehicle planning candidates:
\begin{equation}
    (\mathbf{M}_{e}, \mathbf{P}_{e}) = \Psi_{e}(\mathbf{H}_{e}),
\end{equation}
where $\mathbf{M}_{e}$ denotes the motion prediction output and $\mathbf{P}_{e}$ denotes the planning output of expert $e$. Importantly, the experts are not connected by cross-expert attention during representation learning. This prevents semantic leakage among branches and preserves the independence of different planning factors. Compared with conventional mixture-of-experts designs that usually place experts inside generic feed-forward layers, PrismAD treats each expert as a complete motion-planning branch. Therefore, the specialization occurs at the planning level rather than at a low-level feature transformation stage.

\subsection{Semantics-Aware Router and Expert Aggregation}
Although semantic partitioning enables expert specialization, different driving scenarios require different planning factors. For example, interaction reasoning is more important in dense traffic, geometry reasoning is more critical near intersections or curved roads, and intent reasoning becomes essential when the driving command/status changes. Therefore, PrismAD introduces a semantics-aware router to adaptively evaluate and aggregate the expert outputs.

The router takes the ego feature, driving command/status, detection tokens, and map tokens as inputs. The driving command/status is first projected into the token feature space and concatenated with the ego token to form an ego-centric query. The detection and map tokens are concatenated as key/value tokens. A cross-attention module is then used to aggregate decision-relevant scene context:
\begin{equation}
    \mathbf{r} = \mathrm{RouterAttn}([\mathbf{x}_{\mathrm{ego}}; \mathbf{x}_{\mathrm{cmd}}], 
    [\mathcal{T}_{\mathrm{det}}; \mathcal{T}_{\mathrm{map}}]),
\end{equation}
where $\mathbf{r}$ is the router feature. This design allows the router to estimate expert importance from an ego-centric perspective while still considering agents and map structures.

The router feature $\mathbf{r}$ is then fed into lightweight MLPs to produce routing logits:
\begin{equation}
    \mathbf{z} = \mathrm{MLP}_{r}(\mathbf{r}) \in \mathbb{R}^{6}.
\end{equation}
The first three logits mean the motion experts, while the last three logits correspond to the planning experts:
\begin{equation}
    \mathbf{z}^{m} = \mathbf{z}_{1:3}, 
    \quad 
    \mathbf{z}^{p} = \mathbf{z}_{4:6}.
\end{equation}
This decoupled routing design lets PrismAD assign different expert preferences for motion prediction and ego planning. During training, we inject Gaussian noise into routing logits to encourage exploration and reduce expert collapse:
\begin{equation}
    \tilde{\mathbf{z}} = \mathbf{z} + \epsilon \odot \sigma(\mathbf{r}), 
    \quad \epsilon \sim \mathcal{N}(0, I),
\end{equation}
where $\sigma(\mathbf{r})$ is the predicted noise scale. Top-$k$ selection and softmax normalization are then independently applied to the motion and planning logits:
\begin{equation}
\begin{aligned}
\mathbf{w}^{m} &= \mathrm{Softmax}\!\left(\mathrm{TopK}(\tilde{\mathbf{z}}^{m})\right), \\
\mathbf{w}^{p} &= \mathrm{Softmax}\!\left(\mathrm{TopK}(\tilde{\mathbf{z}}^{p})\right).
\end{aligned}
\end{equation}
The obtained weights are used for both sparse expert activation and output aggregation. The interaction expert is always activated as the basic planning branch, while the geometry and intent experts are selectively activated according to the routing scores. In practice, we keep the interaction expert active for all samples because it contains temporally propagated agent information, which plays a critical role in motion prediction, respectively. This enables sample-wise adaptive computation and avoids unnecessary expert execution.

Motion and planning outputs are fused separately:
\begin{equation}
    \mathbf{M} = \sum_{e} w^{m}_{e}\mathbf{M}_{e},
    \quad
    \mathbf{P} = \sum_{e} w^{p}_{e}\mathbf{P}_{e}.
\end{equation}
The fused motion output $\mathbf{M}$ and planning output $\mathbf{P}$ are decoded to produce the final agent trajectories and ego trajectory. Since the router produces factor-specific weights, the final decision can be interpreted by inspecting the contributions of interaction, geometry, and intent experts. This makes PrismAD more transparent than coupled planners and enables adaptive planning in diverse driving scenarios.

\subsection{Expert Specialization and Router Training}

To stabilize the optimization of PrismAD and avoid disrupting the perception and planning capability, we adopt a three-stage training strategy. The training process gradually transfers the model from a standard end-to-end planner to a semantic mixture-of-planners, and finally learns the adaptive routing module for expert fusion.

In the first stage, we train the base end-to-end planning framework following the SparseDrive\cite{Sun2024SparseDriveEA} or DiffusionDrive\cite{diffusiondrive} training protocol. This stage builds a strong perception-planning foundation, including object detection, online mapping and provides reliable scene tokens for the subsequent semantic expert construction. 

In the second stage, we replace the original planning head with the proposed semantic mixture-of-planners and train three independent experts, including the interaction expert, geometry expert, and intent expert. Each expert receives its corresponding semantic tokens and is optimized with its own motion and planning supervision. Specifically, the interaction expert learns agent-centric interaction reasoning from detection tokens, the geometry expert captures road-structure constraints from map tokens, and the intent expert models command-conditioned planning behavior from navigation tokens. At this stage, the router is not activated, and all experts are trained with independent losses. This design encourages each expert to acquire specialized planning capability before being fused, avoiding early-stage competition among untrained experts.


In the third stage, we freeze the perception backbone and all expert branches, and train only the router and fusion modules. The router predicts expert weights for motion prediction and ego planning, and the final loss is computed only on fused outputs. During this stage, we adopt sparse top-$K$ activation with noisy gating to encourage robust expert selection while preserving inference efficiency. Since expert parameters are fixed, the router learns how to combine already-specialized experts rather than changing their internal representations, which isolates the optimization of adaptive expert fusion and improves training stability.

\section{EXPERIMENTS}
\subsection{Datasets and Evaluation Benchmarks}

We conduct experiments on three nuScenes-based benchmarks, including the standard nuScenes\cite{caesar2020nuscenes} validation set, NeuroNCAP\cite{ljungbergh2024neuroncap}, and Turning-nuScenes\cite{Song2025DontST}. These benchmarks evaluate complementary aspects of planning: standard open-loop trajectory prediction, closed-loop safety in safety-critical scenarios, and robustness in turning scenarios.

\textbf{nuScenes.}
We first evaluate open-loop planning performance on the standard nuScenes benchmark. NuScenes is a large-scale real-world autonomous driving dataset with 1,000 driving scenes, each lasting approximately 20 seconds. It provides synchronized data from a full sensor suite, including 6 cameras, 1 LiDAR, 5 radars, GPS, and IMU sensors. Following the common end-to-end planning protocol, we use only multi-view camera images as visual input, together with the required navigation-command information for trajectory prediction. Following~\cite{Li2023IsES}, we do not use ego status as input to avoid shortcut learning.

\textbf{NeuroNCAP.}
To assess closed-loop safety, we further evaluate our method on the NeuroNCAP benchmark. NeuroNCAP constructs safety-critical closed-loop scenarios from nuScenes logs using photorealistic neural rendering. It contains three collision-oriented scenarios, including stationary, frontal, and side scenarios, where the ego vehicle must react properly to avoid or mitigate collisions. Unlike standard open-loop evaluation, NeuroNCAP executes the predicted trajectory through a controller and vehicle model, iteratively rendering new sensor observations for closed-loop testing.

\textbf{Turning-nuScenes.}
We also report results on Turning-nuScenes, a challenging subset curated from the nuScenes validation set. Since many nuScenes planning samples correspond to relatively simple straight-driving cases, Turning-nuScenes focuses on turning scenarios to better examine planning robustness under stronger geometric and intention constraints. Following MomAD\cite{Song2025DontST}, the subset is constructed from turning samples in nuScenes validation and contains 17 scenes with 680 samples. 



\subsection{Evaluation Metrics}

For open-loop planning evaluation on nuScenes and Turning-nuScenes, we report the average L2 displacement error and collision rate as the main metrics. L2 error measures the distance between the predicted ego trajectory and the ground-truth trajectory, following the protocol of VAD~\cite{Jiang2023VADVS}. The collision rate is computed under the evaluation setting used in SparseDrive~\cite{Sun2024SparseDriveEA}, reflecting whether the predicted trajectory leads to collisions with surrounding agents.

For closed-loop safety evaluation on NeuroNCAP, we report the NeuroNCAP score and collision rate. The NeuroNCAP score follows a five-star-style safety protocol, where a full score is assigned when the collision is completely avoided, and a partial score is given according to the reduction of impact velocity when a collision occurs\cite{ljungbergh2024neuroncap}. The collision rate measures the percentage of closed-loop trials that result in collision. Following the official setting, we report the scores and collision rates for each scenario type, as well as the average over all scenarios.

\subsection{Implementation Details}
We implement PrismAD based on the SparseDrive\cite{Sun2024SparseDriveEA} or DiffusionDrive\cite{diffusiondrive} framework. Following the common nuScenes setting, we use ResNet50\cite{He2015DeepRL} as the image backbone and FPN as the neck to extract multi-scale image features from six surrounding cameras. The input image resolution is $704 \times 256$. For planning, the future ego trajectory is predicted over a 3-second horizon with 6 waypoints.

We train the model using AdamW\cite{Loshchilov2017DecoupledWD} with a learning rate of $1\times10^{-4}$ and weight decay of $1\times10^{-3}$. The total batch size is set to 24 on 4 GPUs. We train the expert stage for 24 epochs, followed by 5 epochs of router-only training.

\subsection{Main Results}

\textbf{Open-loop Planning Results.}
Table~\ref{tab:open-loop planning} reports the open-loop planning results on the nuScenes validation set. PrismAD consistently improves two strong baselines, SparseDrive and DiffusionDrive, demonstrating its compatibility with different planning frameworks. Based on SparseDrive, PrismAD reduces the average L2 error from $0.61$ m to $0.59$ m and lowers the average collision rate from $0.08\%$ to $0.07\%$, while maintaining comparable inference speed. The gains are observed across all prediction horizons, indicating that semantic expert decomposition improves trajectory accuracy without sacrificing safety.

When applied to DiffusionDrive, PrismAD achieves stronger safety improvement. The average L2 error decreases from $0.57$ m to $0.55$ m, and the average collision rate is reduced from $0.08\%$ to $0.04\%$. These results show that PrismAD is not tied to a specific backbone planner, but can serve as a general semantic mixture framework for improving both trajectory quality and long-horizon planning safety.

\textbf{Closed-loop Planning Results.}
Table~\ref{tab:closed-loop planning} reports the closed-loop simulation results on the NeuroNCAP benchmark. Compared with SparseDrive, PrismAD$_{SparseDrive}$ improves the average NeuroNCAP score from $0.92$ to $1.83$ and reduces the average collision rate from $93.9\%$ to $82.0\%$, corresponding to a relative improvement of $98.9\%$ and a relative collision reduction of $12.7\%$, respectively.

PrismAD also improves DiffusionDrive in closed-loop evaluation. PrismAD$_{DiffusionDrive}$ increases the average NeuroNCAP score from $3.00$ to $3.28$ and decreases the average collision rate from $51.9\%$ to $44.1\%$, yielding a relative score improvement of $9.3\%$ and relative collision reduction of $15.0\%$. These results show that our semantic mixture design improves the safety of executed plans in closed-loop simulation and generalizes to different base planners.

\begin{table*}[htbp]
  \vspace{2mm}
  \caption{Open-loop planning results on the nuScenes validation dataset. FPS is measured on an NVIDIA Tesla A100 GPU.
  }
  \label{tab:open-loop planning}
  \centering
  \resizebox{0.9\linewidth}{!}{%
  \begin{tabular}{l|c| c c c c|c c c c|c}
    \toprule
    \multirow{2}{*}{\textbf{Method}} & \multirow{2}{*}{\textbf{Reference}}
    & \multicolumn{4}{c|}{\textbf{L2 (m)}$\downarrow$}
    & \multicolumn{4}{c|}{\textbf{Col. Rate (\%)}$\downarrow$}
    & \multirow{2}{*}{\textbf{FPS}$\uparrow$} \\
    & &
    1s & 2s & 3s & \cellcolor{gray!15}Avg.
    & 1s & 2s & 3s & \cellcolor{gray!15}Avg.
    & \\
    \midrule
    \multicolumn{11}{c}{\textbf{VLA \& World Model-based}} \\    
    \midrule
    Drive-WM\cite{Wang2023DrivingIT}& CVPR2023    
    & 0.43 & 0.77 & 1.20 & \cellcolor{gray!15}0.80    
    & 0.10 & 0.21 & 0.48 & \cellcolor{gray!15}0.26    
    & - \\

    LAW\cite{li2024enhancing}& ArXiv2024        
    & \textbf{0.26} & 0.57 & 1.01 & \cellcolor{gray!15}0.61        
    & 0.14 & 0.21 & 0.54 & \cellcolor{gray!15}0.30        
    & - \\

    OccWorld\cite{zheng2024occworld} & ECCV2024        
    & 0.39 & 0.73 & 1.18 & \cellcolor{gray!15}0.77        
    & 0.11 & 0.19 & 0.67 & \cellcolor{gray!15}0.32        
    & - \\

    ORION\cite{fu2025orion} & ICCV2025        
    & 0.40 & 0.80 & 1.32 & \cellcolor{gray!15}0.84        
    & 0.04 & 0.46 & 2.32 & \cellcolor{gray!15}0.94        
    & - \\
    
    OmniDrive\cite{wang2025omnidrive} & CVPR2025        
    & 0.40 & 0.80 & 1.32 & \cellcolor{gray!15}0.84        
    & 0.04 & 0.46 & 2.32 & \cellcolor{gray!15}0.94        
    & - \\

    UniDriveVLA\cite{Li2026UniDriveVLAUU} & ArXiv2026        
    & 0.28 & \textbf{0.51} & \textbf{0.82} & \cellcolor{gray!15}\textbf{0.54}        
    & 0.08 & 0.13 & 0.31 & \cellcolor{gray!15}0.17        
    & - \\    
    \midrule
    \multicolumn{11}{c}{\textbf{End-to-End Planning}} \\
    \midrule
    VAD\cite{Jiang2023VADVS}         & ICCV2023
    & 0.41 & 0.70 & 1.05 & \cellcolor{gray!15}0.72
    & 0.03 & 0.19 & 0.43 & \cellcolor{gray!15}0.21
    & 3.9 \\

    UniAD\cite{hu2023_uniad}         & CVPR2023    
    & 0.45 & 0.70 & 1.04 & \cellcolor{gray!15}0.73    
    & 0.62 & 0.58 & 0.63 & \cellcolor{gray!15}0.61    
    & 3.9 \\
    
    MomAD\cite{Song2025DontST}        & CVPR2025
    & 0.31 & 0.57 & 0.91 & \cellcolor{gray!15}0.60
    & \underline{0.01} & 0.05 & 0.22 & \cellcolor{gray!15}0.09
    & 5.8 \\

    UncAD\cite{Yang2025UncADTS} & ICRA2025            
    & 0.29 & 0.57 & 0.95 & \cellcolor{gray!15}0.60            
    & \textbf{0.00} & \underline{0.04} & 0.18 & \cellcolor{gray!15}\underline{0.07}            
    & - \\

    
    GenAD\cite{Zheng2024GenADGE}      & ECCV2024
    & 0.36 & 0.83 & 1.55 & \cellcolor{gray!15}0.91
    & 0.06 & 0.23 & 1.00 & \cellcolor{gray!15}0.43
    & 5.0 \\

    BridgeAD\cite{Zhang2025BridgingPA} & CVPR2025    
    & 0.29 & 0.57 & 0.92 & \cellcolor{gray!15}0.59    
    & \underline{0.01} & 0.05 & 0.22 & \cellcolor{gray!15}0.09    
    & 3.7 \\

    BEV-Planner\cite{Li2023IsES} & CVPR2024        
    & 0.30 & \textbf{0.52} & \underline{0.83} & \cellcolor{gray!15}\underline{0.55}        
    & 0.10 & 0.37 & 1.30 & \cellcolor{gray!15}0.59        
    & - \\

    CausalVAD\cite{Tang2026CausalVADDE} & CVPR2026        
    & \underline{0.27} & \underline{0.52} & \textbf{0.82} & \cellcolor{gray!15}\textbf{0.54}        
    & 0.02 & 0.09 & 0.22 & \cellcolor{gray!15}0.11        
    & 5.5 \\
    
    FlowAD\cite{FlowAD2026} & ICLR2026         
    & - & - & - & \cellcolor{gray!15}\textbf{0.54}            
    & - & - & - & \cellcolor{gray!15}\textbf{0.04}           
    & 4.0 \\    
    \midrule
    SparseDrive\cite{Sun2024SparseDriveEA} & ICRA2025
    & 0.29 & 0.58 & 0.96 & \cellcolor{gray!15}0.61
    & \underline{0.01} & 0.05 & 0.18 & \cellcolor{gray!15}0.08
    & \textbf{6.7} \\

    \rowcolor{gray!15}
    PrismAD$_{SparseDrive}$ & Ours    
    & 0.28 & 0.56 & 0.94 & \cellcolor{gray!15}0.59\textcolor{magenta}{$\downarrow$3.3\%}   
    & \textbf{0.00} & \underline{0.04} & 0.17 & \cellcolor{gray!15}\underline{0.07}\textcolor{magenta}{$\downarrow$12.5\%}       
    &  \underline{6.3} \\

    DiffusionDrive\cite{diffusiondrive} & CVPR2025    
    & \underline{0.27} & 0.54 & 0.90 & \cellcolor{gray!15}0.57    
    & 0.03 & 0.05 & \underline{0.16} & \cellcolor{gray!15}0.08        
    & 5.1 \\

    \rowcolor{gray!15}
    PrismAD$_{DiffusionDrive}$ & Ours    
    & \underline{0.27} & \underline{0.52} & 0.86 & \cellcolor{gray!15}\underline{0.55}\textcolor{magenta}{$\downarrow$3.6\%}    
    & \underline{0.01} & \textbf{0.02} & \textbf{0.10} & \cellcolor{gray!15}\textbf{0.04}\textcolor{magenta}{$\downarrow$50.0\%}  
    &  5.5 \\
  \bottomrule
  \end{tabular}%
  }
  \vspace{-3mm}
\end{table*}

\begin{table}[htbp]
    \caption{Closed-loop simulation results on nuScenes dataset with NeuroNCAP benchmark. $^*$ refers the use of official weights.}
    \label{tab:closed-loop planning}
    \centering
    \resizebox{\linewidth}{!}{
    \begin{tabular}{@{}l c c c c c c c c}
        \toprule
        \multirow{2}{*}{Method} & \multicolumn{4}{c}{NeuroNCAP scores$\uparrow$} & \multicolumn{4}{c}{Collision rates(\%)$\downarrow$} \\
         & Stat. & Frontal & Side & \cellcolor{gray!15}Avg. & Stat. & Frontal & Side & \cellcolor{gray!15}Avg. \\
         \midrule
         Impromptu-VLA\cite{Chi2025ImpromptuVO} & 1.77 & 2.31 & 2.10 & \cellcolor{gray!15}2.15 & 70.0 & 59.0 & 65.0 & \cellcolor{gray!15}65.5 \\
         Reasoning-VLA-3B\cite{Zhang2025ReasoningVLAAF} & 1.88 & 2.29 & 1.94 & \cellcolor{gray!15}2.04 & 63.7 & 60.4 & 64.1 & \cellcolor{gray!15}62.7 \\
         
         Reasoning-VLA-7B\cite{Zhang2025ReasoningVLAAF} & 1.93 & \textbf{2.57} & 2.24 & \cellcolor{gray!15}2.25 & 59.8 & \textbf{56.0} & 62.2 & \cellcolor{gray!15}59.4 \\

         Reasoning-VLA-7B+\cite{Zhang2025ReasoningVLAAF} & 2.06 & 2.33 & 2.17 & \cellcolor{gray!15}2.19 & 57.9 & 57.4 & 64.0 & \cellcolor{gray!15}59.8 \\
         \midrule
         UniAD\cite{hu2023_uniad} & 0.84 & 0.10 & 1.26 & \cellcolor{gray!15}0.73 & 87.8 & 98.4 & 79.6 & \cellcolor{gray!15}88.6 \\
         VAD\cite{Jiang2023VADVS} & 0.47 & 0.04 & 1.45 & \cellcolor{gray!15}0.66 & 96.2 & 99.6 & 81.6 & \cellcolor{gray!15}92.5 \\
        SparseDrive\cite{Sun2024SparseDriveEA} & - & - & - & \cellcolor{gray!15}0.92 & - & - & - & \cellcolor{gray!15}93.9\\
         BridgeAD-S\cite{Zhang2025BridgingPA} & - & - & - & \cellcolor{gray!15}1.52 & - & - & - & \cellcolor{gray!15}76.2\\

         BridgeAD-B\cite{Zhang2025BridgingPA} & - & - & - & \cellcolor{gray!15}1.60 & - & - & - & \cellcolor{gray!15}72.6\\
         DiffusionDrive$^*$\cite{diffusiondrive} & 3.88 & 2.37 & 2.74 & \cellcolor{gray!15}3.00 & 32.4 & 71.6 & 51.6 & \cellcolor{gray!15}51.9 \\
         \midrule
         \textbf{PrismAD}$_{SparseDrive}$ & 2.50 & 1.57 & 1.41 & \cellcolor{gray!15}1.83\textcolor{magenta}{$\uparrow$98.9\%}  & 72.4 & 89.6 & 84.0 & \cellcolor{gray!15}82.0\textcolor{magenta}{$\downarrow$12.7\%}  \\
         \textbf{PrismAD}$_{DiffusionDrive}$ & \textbf{4.31} & 2.28 & \textbf{3.25} & \cellcolor{gray!15}\textbf{3.28}\textcolor{magenta}{$\uparrow$9.3\%}  & \textbf{21.8} & 75.2 & \textbf{35.2} & \cellcolor{gray!15}\textbf{44.1}\textcolor{magenta}{$\downarrow$15.0\%}  \\
    \bottomrule
    \end{tabular}
    }
\end{table}
\textbf{Planning Results on Turning-nuScenes.}
Table~\ref{tab:Turning planning} evaluates planning performance on the challenging Turning-nuScenes benchmark. PrismAD demonstrably enhances the baseline planners, SparseDrive and DiffusionDrive, in both trajectory fidelity and safety metrics. For the SparseDrive baseline, PrismAD reduces the average L2 error from $0.86$ m to $0.71$ m and lowers the average collision rate from $0.40\%$ to $0.07\%$.
For the DiffusionDrive baseline, PrismAD further achieves the best overall performance, reducing the average L2 error from $0.62$ m to $0.59$ m and the average collision rate from $0.06\%$ to $0.01\%$. The improvement is most evident in long-horizon safety, where the 3s collision rate decreases from $0.06\%$ to $0.03\%$. These comparisons indicate that semantic expert decomposition is especially effective in command-sensitive turning scenarios.
\begin{table}[htbp]
    \caption{Planning results on the Turning-nuScenes\cite{Song2025DontST} validation dataset. $^*$ refers the use of official weights.}
    \label{tab:Turning planning}
    \centering
    \resizebox{\linewidth}{!}{
    \begin{tabular}{@{}l c c c c c c c c}
        \toprule
        \multirow{2}{*}{Method} & \multicolumn{4}{c}{L2 (m)$\downarrow$} & \multicolumn{4}{c}{Col. Rate(\%)$\downarrow$} \\
         & 1s & 2s & 3s & \cellcolor{gray!15}Avg. & 1s & 2s & 3s & \cellcolor{gray!15}Avg. \\
         \midrule
         SparseDrive\cite{Sun2024SparseDriveEA} & 0.35 & 0.77 & 1.46 & \cellcolor{gray!15}0.86 & 0.04 & 0.17 & 0.98 & \cellcolor{gray!15}0.40 \\
         MomAD\cite{Song2025DontST} & 0.33 & 0.70 & 1.24 & \cellcolor{gray!15}0.76 & 0.03 & 0.13 & 0.79 & \cellcolor{gray!15}0.32 \\
         DiffusionDrive$^*$\cite{diffusiondrive} & 0.29 & 0.59 & 0.98 & \cellcolor{gray!15}0.62 & 0.09 & 0.04 & 0.06 & \cellcolor{gray!15}0.06 \\
         \textbf{PrismAD}$_{SparseDrive}$ & 0.31 & 0.66 & 1.15 & \cellcolor{gray!15}0.71\textcolor{magenta}{$\downarrow$17.4\%} & \textbf{0.00} & 0.04 & 0.17 & \cellcolor{gray!15}0.07\textcolor{magenta}{$\downarrow$82.5\%} \\
         \textbf{PrismAD}$_{DiffusionDrive}$ & \textbf{0.28} & \textbf{0.56} & \textbf{0.93} & \cellcolor{gray!15}\textbf{0.59}\textcolor{magenta}{$\downarrow$4.8\%} & \textbf{0.00} & \textbf{0.00} & \textbf{0.03} & \cellcolor{gray!15}\textbf{0.01}\textcolor{magenta}{$\downarrow$83.3\%} \\
    \bottomrule
    \end{tabular}
    }
    \vspace{-2mm}
\end{table}

\begin{table}[tbp]
    \caption{Ablation on Semantic Experts.}
    \label{tab:Ablation on Semantic Experts}
    \centering
    \resizebox{\linewidth}{!}{
    \begin{tabular}{c c c|c c}
        \toprule
        \textbf{Interaction} & 
        \textbf{Geometry} & \textbf{Intent} & \textbf{Avg. L2}$\downarrow$ & \textbf{Avg. Col. Rate}$\downarrow$ \\
         \midrule
         \cmark & \xmark & \xmark & 0.552 & 0.064 \\
         \cmark & \cmark & \xmark & 0.552 & 0.054 \\
         \cmark & \xmark & \cmark & 0.553 & 0.052 \\
         \cmark & \cmark & \cmark & \textbf{0.550} & \textbf{0.042} \\
         \bottomrule
    \end{tabular}
    }
    \vspace{-2mm}
\end{table}

\subsection{Ablation Study}
\textbf{Ablation on Semantic Experts.}
Table~\ref{tab:Ablation on Semantic Experts} evaluates different semantic expert combinations using the same trained MoE checkpoint, where inactive experts are masked during router fusion. This setting isolates the effect of expert composition while keeping the backbone, training schedule, and expert parameters unchanged.

The interaction-only variant provides a strong accuracy baseline with an average L2 error of $0.552$ m, but its average collision rate remains relatively high at $0.064\%$. Adding the geometry expert preserves the same average L2 error and reduces the collision rate to $0.054\%$, while adding the intent expert obtains a slightly higher L2 error of $0.553$ m but further lowers the collision rate to $0.052\%$. Using all three experts achieves the lowest collision rate of $0.042\%$, with only a marginal change in average L2 error compared with the interaction-only baseline. This indicates that geometry and intent do not mainly improve displacement accuracy when used individually, but they provide complementary safety cues when jointly fused with interaction reasoning.

\textbf{Ablation on Router and Fusion.}
Table~\ref{tab:Ablation on router and fusion} studies the effect of different fusion strategies. Uniform Fusion, which averages the outputs of all experts without learned routing, performs the worst, with an average L2 error of $0.62$ m and an average collision rate of $0.12\%$. This shows that simply ensembling semantic experts is insufficient, since different experts should contribute differently across samples.

Introducing learned gating substantially improves performance, reducing the average L2 error to $0.55$ m and the collision rate to $0.05\%$. With noisy gating, the average L2 error remains at $0.55$ m, while the collision rate further decreases to $0.04\%$. This indicates that adaptive router fusion is essential for exploiting specialized experts, and noisy gating further improves planning safety by encouraging more robust expert selection.
\begin{figure*}[t]
\vspace*{2mm}
\centering
\begin{minipage}{0.94\textwidth}
\centering
\begin{minipage}{0.235\textwidth}
\centering
\includegraphics[width=\linewidth]{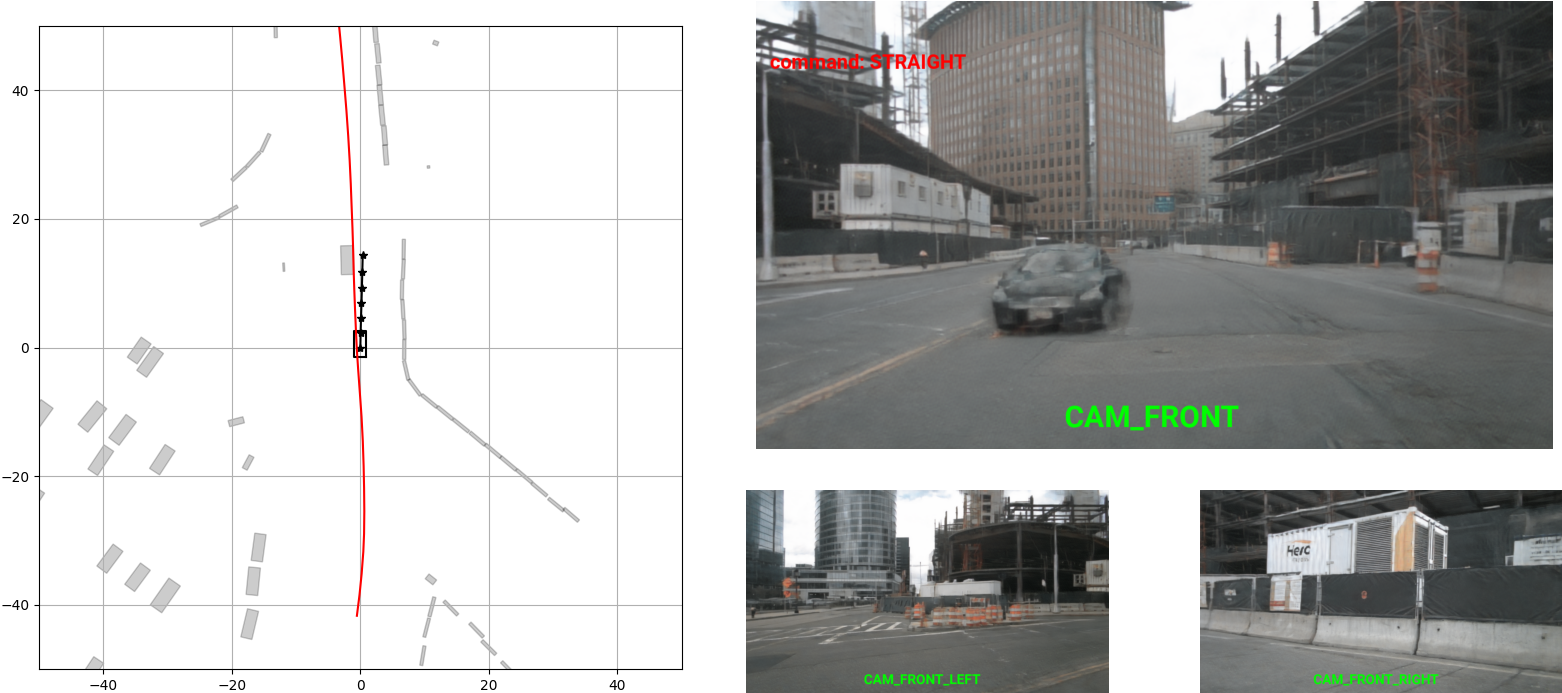}\\[-0.6em]
{\footnotesize $T=1$}
\end{minipage}
\begin{minipage}{0.235\textwidth}
\centering
\includegraphics[width=\linewidth]{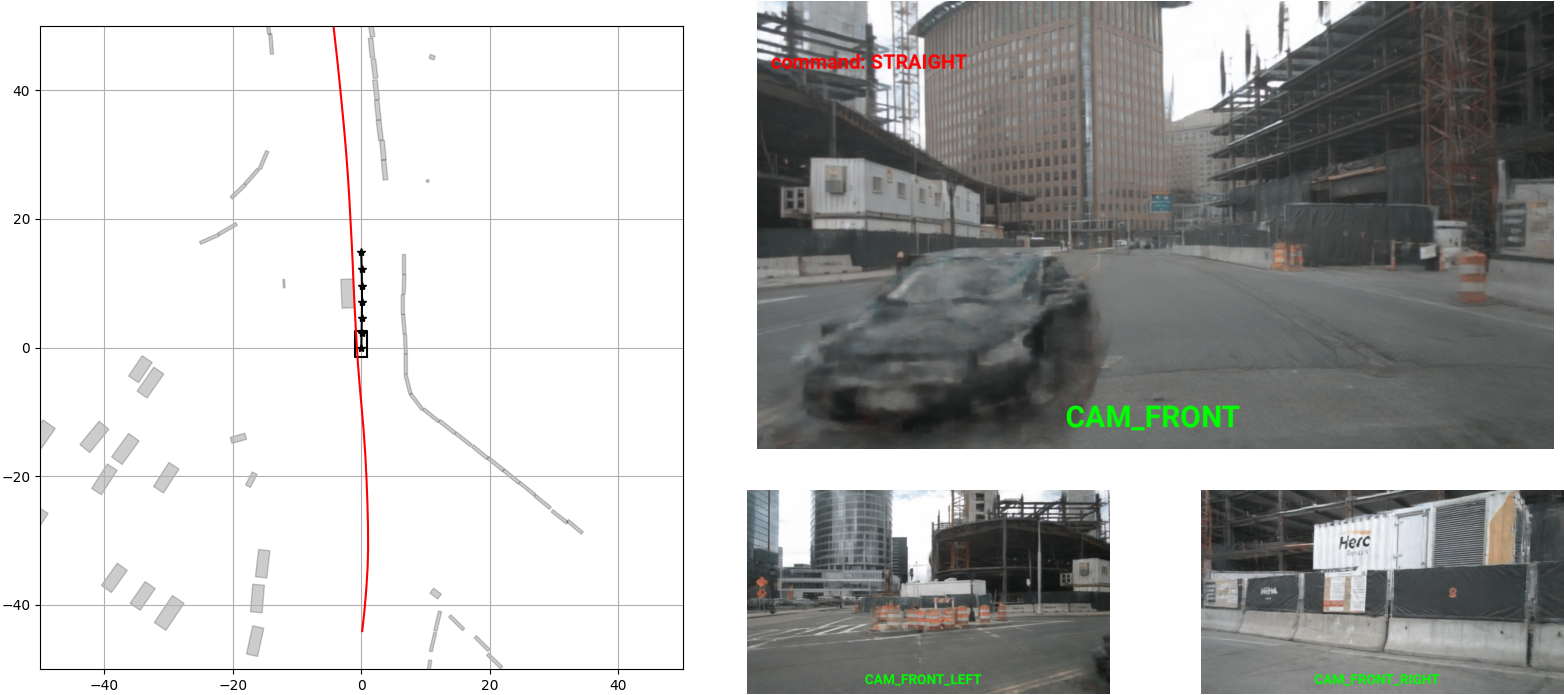}\\[-0.6em]
{\footnotesize $T=2$}
\end{minipage}
\begin{minipage}{0.235\textwidth}
\centering
\includegraphics[width=\linewidth]{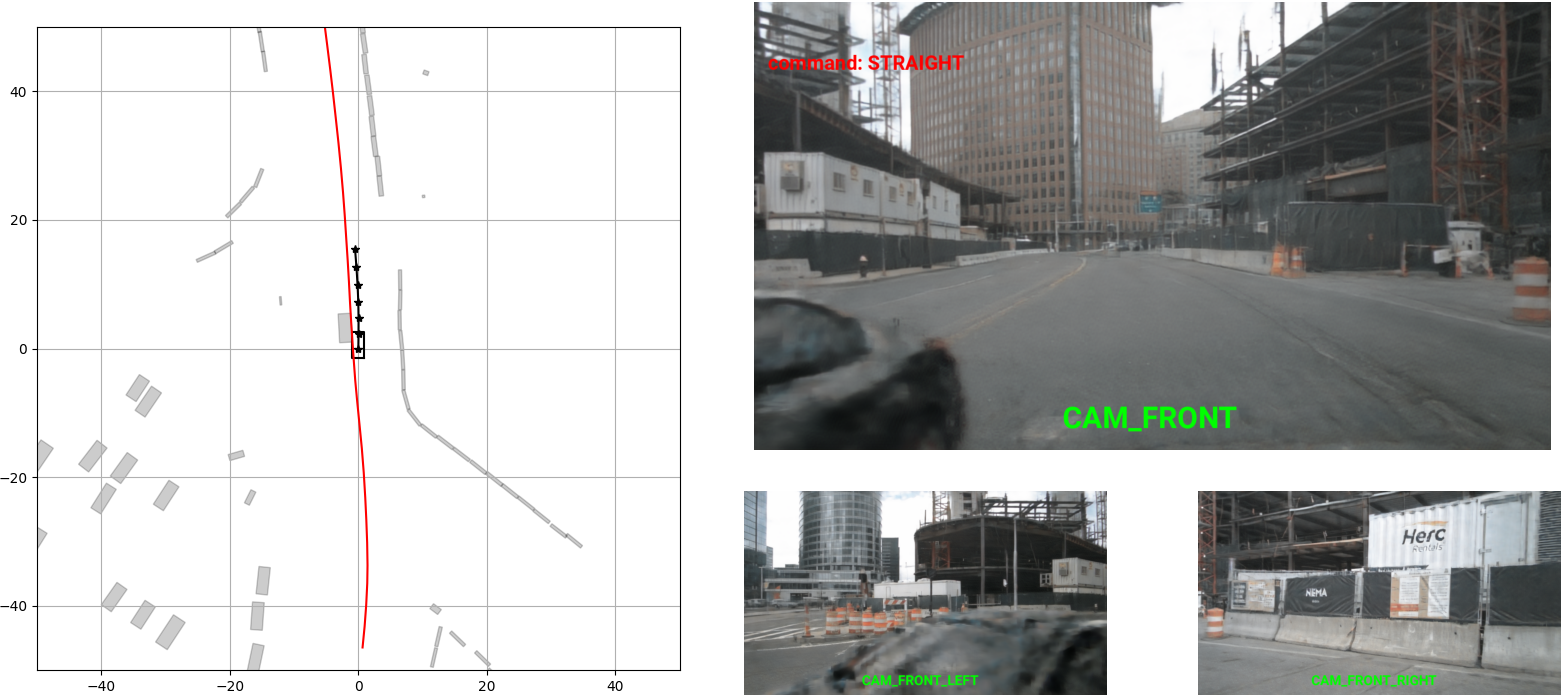}\\[-0.6em]
{\footnotesize $T=3$}
\end{minipage}
\begin{minipage}{0.235\textwidth}
\centering
\includegraphics[width=\linewidth]{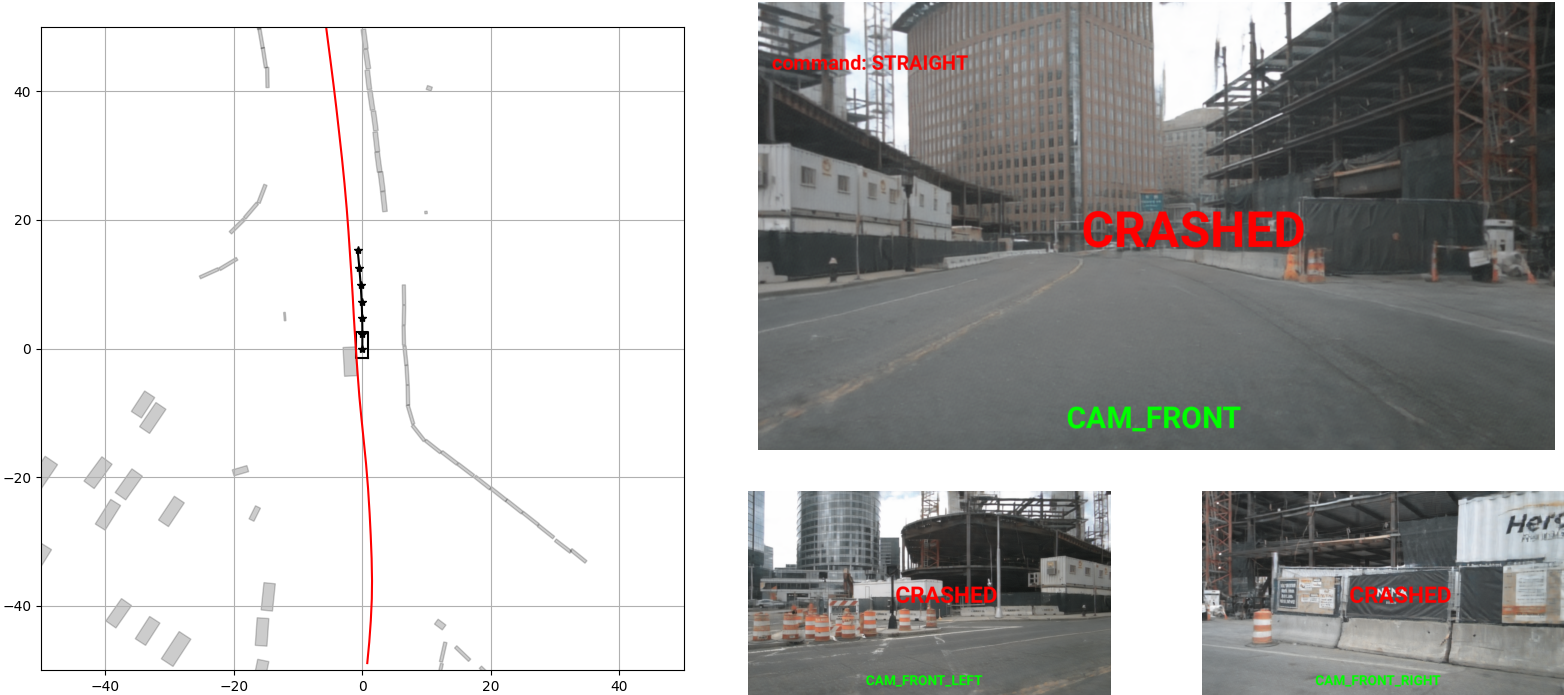}\\[-0.6em]
{\footnotesize $T=4$}
\end{minipage}

{\footnotesize DiffusionDrive}
\end{minipage}

\vspace{0.5em}

\begin{minipage}{0.94\textwidth}
\centering
\begin{minipage}{0.235\textwidth}
\centering
\includegraphics[width=\linewidth]{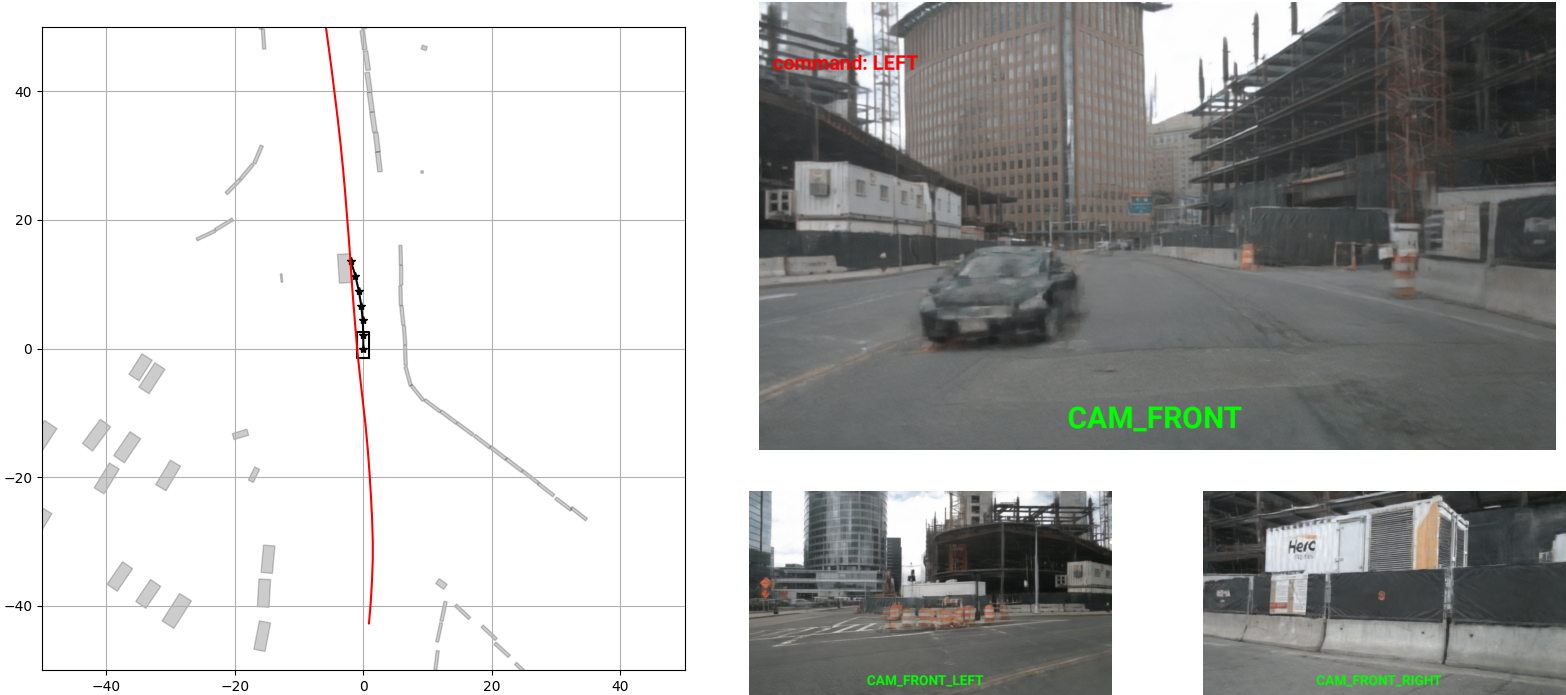}\\[-0.6em]
{\footnotesize $T=1$}
\end{minipage}
\begin{minipage}{0.235\textwidth}
\centering
\includegraphics[width=\linewidth]{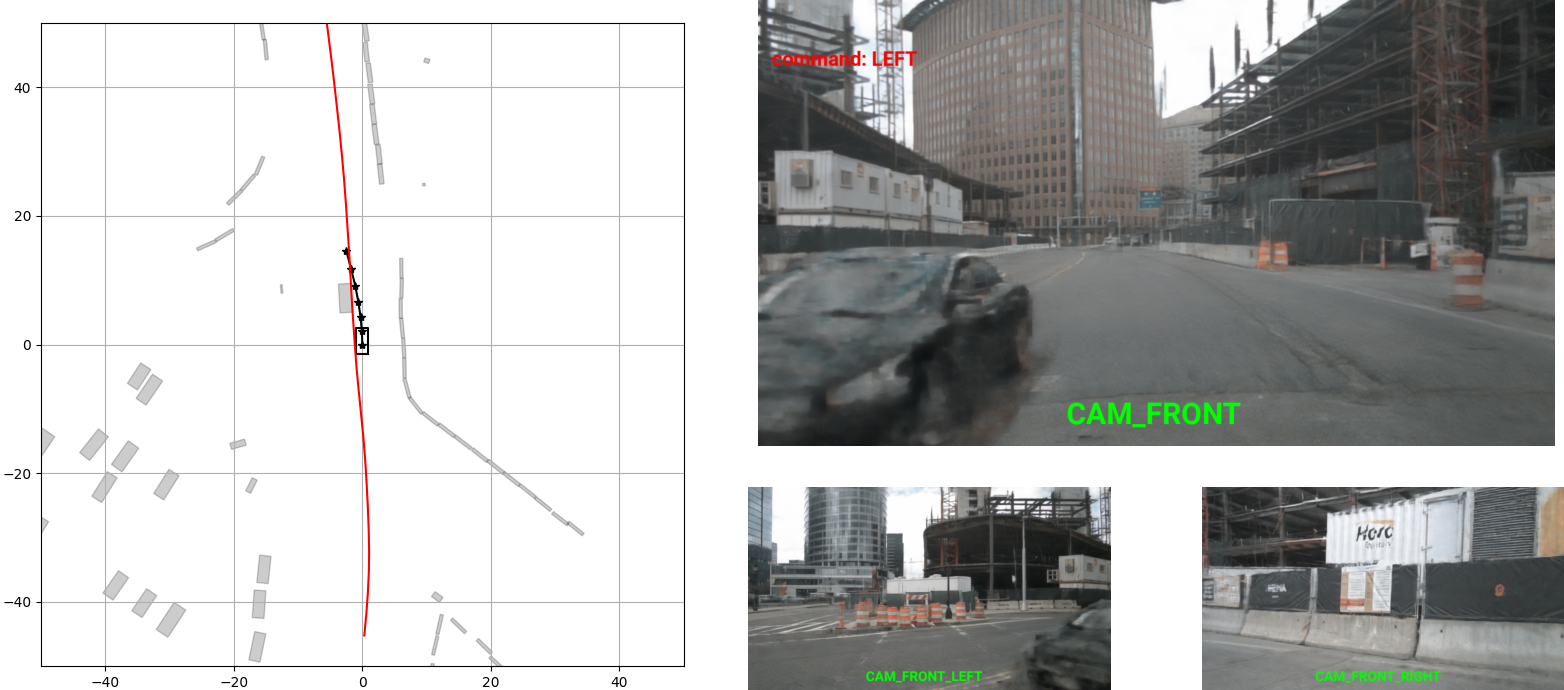}\\[-0.6em]
{\footnotesize $T=2$}
\end{minipage}
\begin{minipage}{0.235\textwidth}
\centering
\includegraphics[width=\linewidth]{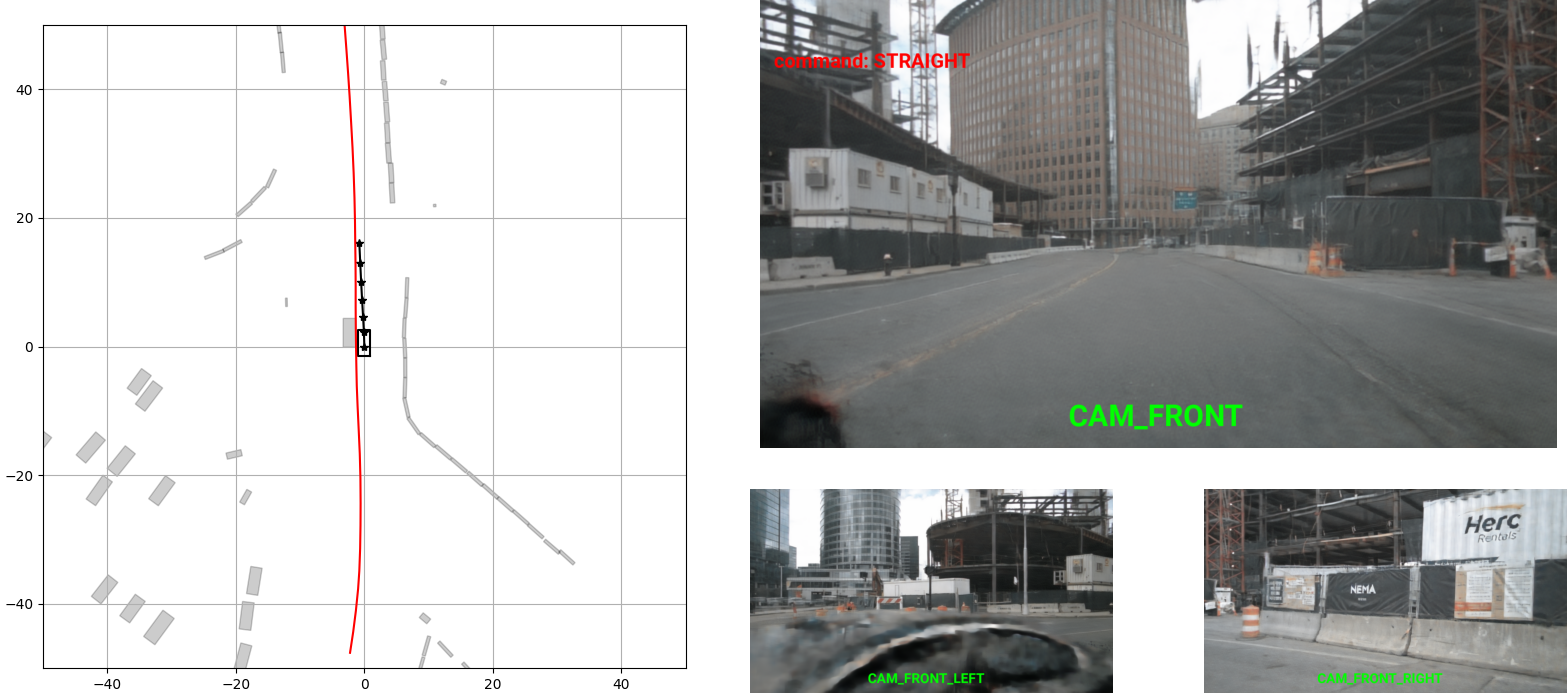}\\[-0.6em]
{\footnotesize $T=3$}
\end{minipage}
\begin{minipage}{0.235\textwidth}
\centering
\includegraphics[width=\linewidth]{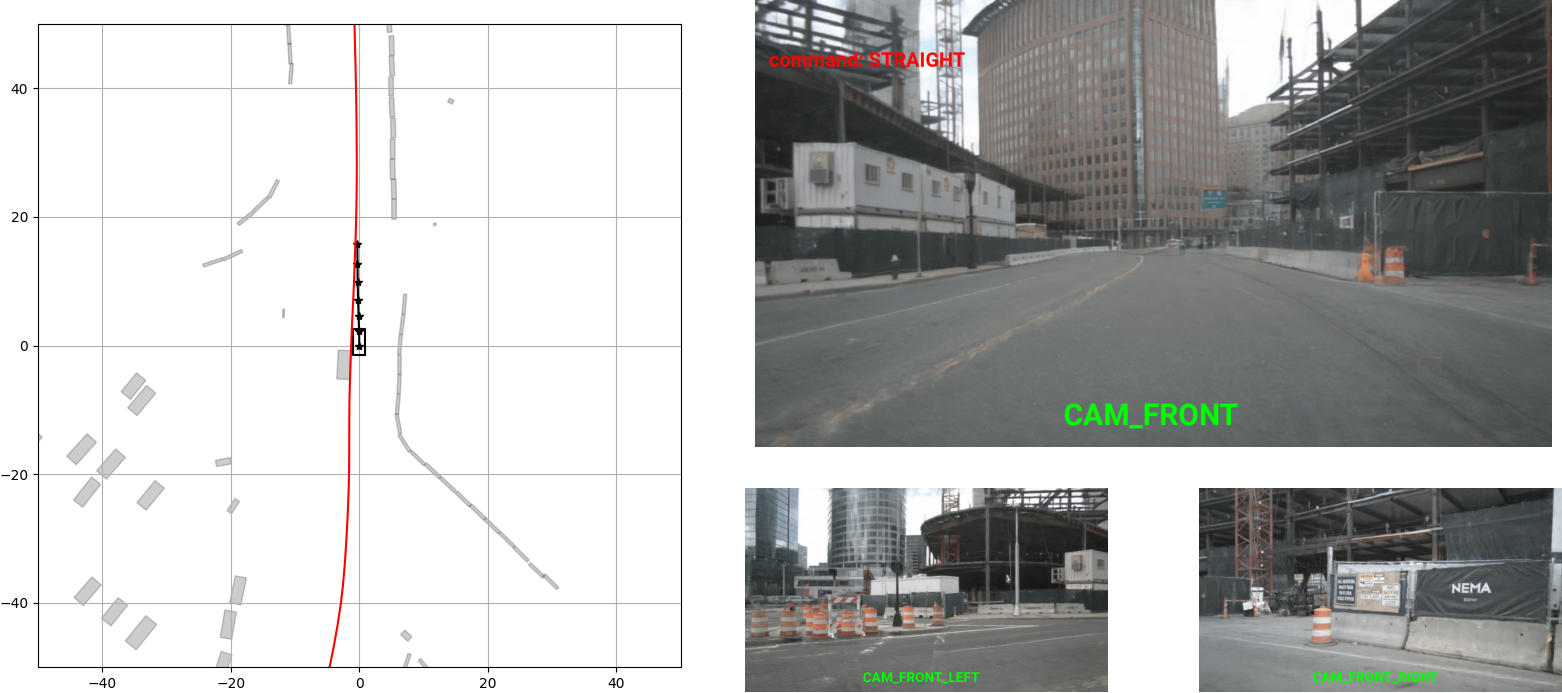}\\[-0.6em]
{\footnotesize $T=4$}
\end{minipage}

{\footnotesize PrismAD (Ours)}
\end{minipage}

\caption{Qualitative analysis results in the closed-loop evaluation.}
\label{fig:qualitative_analysis}
\end{figure*}
\begin{figure*}[tbp]        
    \centering        
    \includegraphics[width=0.9\textwidth]{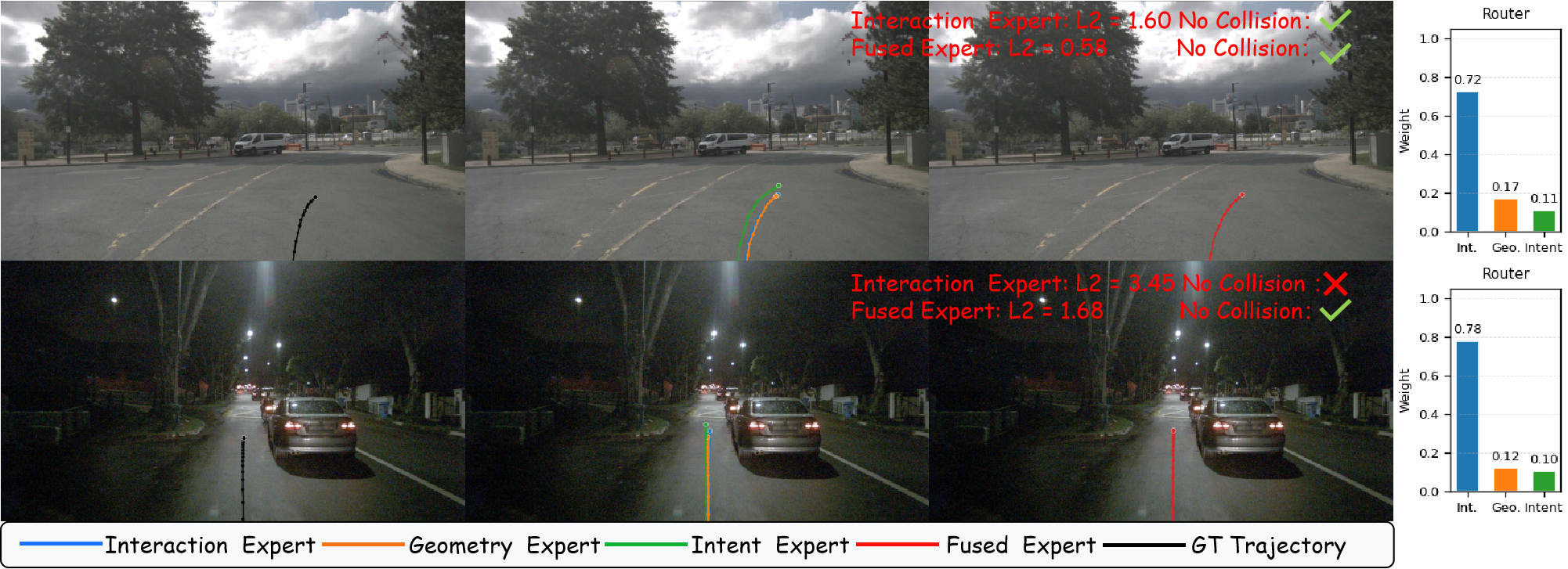} 
    \caption{Qualitative analysis results of PrismAD on the nuScenes.}      
    \label{fig:open-loop}
\end{figure*}

\textbf{Ablation on Sparse Activation.}
Table~\ref{tab:Ablation on sparse activation} evaluates the effect of different top-$K$ settings during inference. Top-1 activation achieves the highest FPS of $6.4$, but its planning performance drops notably, with an average L2 error of $0.74$ m and an average collision rate of $0.17\%$. Increasing the activation budget to Top-2 substantially improves both accuracy and safety, reducing the average L2 error to $0.55$ m and the collision rate to $0.05\%$, while still maintaining $6.0$ FPS. Further increasing to Top-3 keeps the same average L2 error and slightly reduces the collision rate to $0.04\%$, but lowers the FPS to $5.5$. These results indicate that Top-2 provides a favorable trade-off between planning quality and efficiency, while Top-3 offers marginal safety improvement at the cost of additional computation.

\begin{table}[tbp]
    \caption{Ablation on router and fusion.}
    \label{tab:Ablation on router and fusion}
    \centering
    \resizebox{\linewidth}{!}{
    \begin{tabular}{l|l|c|c}
        \toprule
        \textbf{ID} & \textbf{Method} & 
        \textbf{Avg. L2}$\downarrow$ &
        \textbf{Avg. Col. Rate$\downarrow$} \\ 
         \midrule
         \ding{172} & Uniform Fusion & 0.62 & 0.12  \\
         \ding{173} & Gating & 0.55 & 0.05  \\
         \ding{174} & Noisy Gating(Ours) & 0.55 & 0.04  \\
         \bottomrule
    \end{tabular}
    }
\end{table}
\begin{table}[tbp]
    \caption{Ablation on sparse activation.}
    \label{tab:Ablation on sparse activation}
    \centering
    \resizebox{\linewidth}{!}{
    \begin{tabular}{c|c|c|c|c}
        \toprule
        \textbf{Top-$K$} & 
        \textbf{Avg. L2}$\downarrow$ &
        \textbf{Avg. Col. Rate$\downarrow$} &
        \textbf{Params}$\downarrow$ &
        \textbf{FPS}$\uparrow$\\ 
         \midrule
         Top-1 & 0.74 & 0.17 & 89M & 6.4 \\
         Top-2 & 0.55 & 0.05 & 101M & 6.0 \\
         Top-3 & 0.55 & 0.04 & 112M & 5.5 \\
         \bottomrule
    \end{tabular}
    }
    \vspace{-2mm}
\end{table}

\textbf{Qualitative Results.} To provide a comprehensive and intuitive evaluation of our method, we conduct visualization experiments under both open-loop and closed-loop settings. As shown in Fig. \ref{fig:qualitative_analysis}, we compare the real-time obstacle avoidance capability of DiffusionDrive and our method in a frontal scenario across consecutive timesteps. Our model successfully avoids the approaching vehicle. As shown in Fig. \ref{fig:open-loop}, we further compare the planning behaviors of the interaction-only expert and the mixture-of-experts, along with the corresponding expert weights. The results show that although the geography and intent experts are assigned relatively small weights, they remain important for calibrating the trajectory produced by the single interaction expert.

\section{CONCLUSION}

This letter introduces PrismAD, a decoupled end-to-end autonomous driving framework based on a Semantic Mixture-of-Planners. In contrast to conventional planning paradigms that process heterogeneous scene tokens in a coupled representation space, PrismAD explicitly partitions scene tokens into interaction, geometry, and intent groups, and assigns them to independent motion-planning experts. Through its semantics-aware routing mechanism, PrismAD adaptively aggregates expert predictions with separate weights for motion prediction and ego planning, enabling factor-aware decision fusion under diverse driving scenarios. Extensive evaluations on nuScenes, NeuroNCAP, and Turning-nuScenes demonstrate that PrismAD consistently improves strong planning baselines in open-loop accuracy, closed-loop safety, and turning-scenario robustness. These results indicate that semantic decoupling and planning-level expert specialization provide a promising direction for interpretable and robust end-to-end autonomous driving.




\bibliographystyle{IEEEtran}
\bibliography{IEEEabrv,IEEEexample}

\end{document}